\documentclass[conference]{IEEEtran}
\IEEEoverridecommandlockouts
\usepackage{cite}
\usepackage{amsmath,amssymb,amsfonts}
\usepackage{algorithmic}
\usepackage{graphicx}
\usepackage{textcomp}
\usepackage{xcolor}
\usepackage{multirow}
\usepackage{enumitem}
\usepackage{booktabs}
\usepackage{listings}
\usepackage{float}
\usepackage{makecell}
\usepackage[most]{tcolorbox}

\def\BibTeX{{\rm B\kern-.05em{\sc i\kern-.025em b}\kern-.08em
    T\kern-.1667em\lower.7ex\hbox{E}\kern-.125emX}}
\begin{document}

\title{Beyond Sequences: Distilling Structured Decision Memory for LLM Recommendation\\
}


\author{
\IEEEauthorblockN{
{Leikun Liang}\textsuperscript{*}\thanks{* Equal contribution, \dag Corresponding author.}, 
{Guoshuai Wang}\textsuperscript{*}, 
{Xingsheng He}\textsuperscript{\dag}, 
{Yushan Han}, 
{Yunyi Xuan}, 
{Xiaoxiao Xu},
{Lin Qu}
}
\IEEEauthorblockA{
\textit{{Alibaba Group}} \\
{Hangzhou, China} \\
\{{liangleikun.llk, guoshui.wgs, xingsheng.hxs, hanyushan.hys, xuanyunyi.xyy, xiaoxiao.xuxx}\}@{alibaba-inc.com} \\
\{{xide.ql}\}@{taobao.com}
\\
}
}


\maketitle

\begin{abstract}
Despite the adoption of large language models (LLMs) in recommendation systems, prevailing approaches mostly model single-type behaviors (e.g., views or purchases). Even when incorporating multiple behaviors, existing methods flatten heterogeneous actions into homogeneous token sequences, ignoring their distinct decision-making roles. This flattening fails to capture semantic hierarchies and contextual nuances in complex decision-making, such as trade-offs between price and quality. Consequently, performance degrades in critical ``difficult-choice'' scenarios involving highly similar items. 
To bridge this gap, we propose MARI (Memory-Augmented Recommendation with Interpretability), which grounds predictions in explicit, structured decision evidence. MARI maintains a Decision Memory Bank (DMB) that archives users' past rationales as Structured Decision Memories (SDMs): concise records of goals, constraints, and trade-offs. These SDMs are generated offline via Post-Hoc Decision Distillation from heterogeneous behaviors and user-generated content. By retrieving relevant SDMs to augment LLM reasoning, MARI achieves interpretability and scalability without the prohibitive cost of processing long raw sequences. 
Extensive experiments show MARI significantly outperforms state-of-the-art baselines on standard next-item prediction and a newly introduced Difficult Choice Prediction task, incurring low latency overhead by decoupling memory construction from online inference.  Qualitative analyses reveal actionable, human-readable insights into user decision-making, marking a concrete step toward reasoning-aware recommendation systems.
\end{abstract}


\section{Introduction}

\begin{figure}[t]
  \centering
  \includegraphics[width=0.488\textwidth]{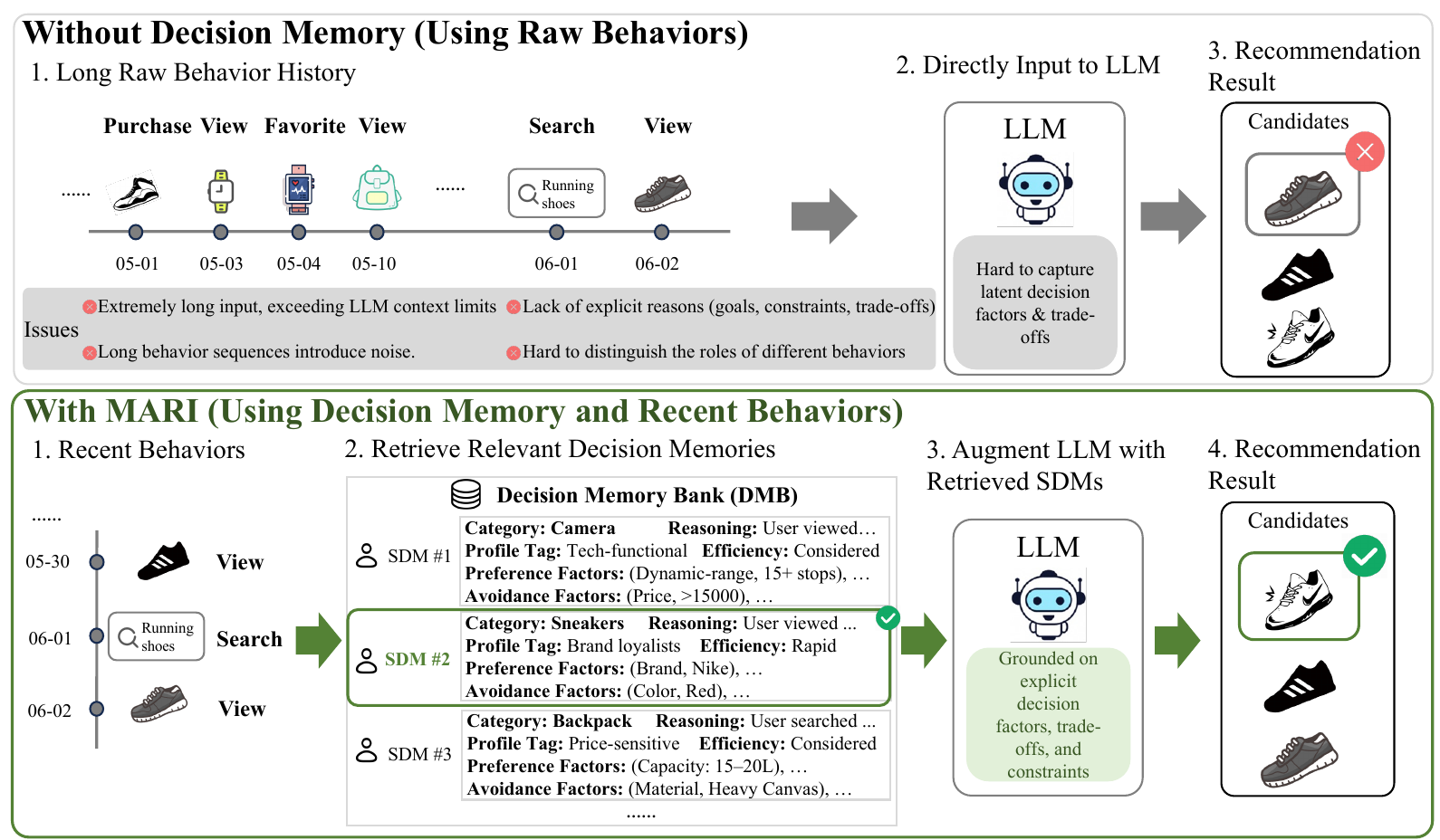}
 \caption{Workflow comparison. \textit{Top:} Directly feeding long raw behavior sequences into an LLM introduces context limits, noise, and latent decision factors, hindering accurate reasoning. \textit{Bottom:} MARI retrieves Structured Decision Memories (SDMs) from a Decision Memory Bank (DMB) based on recent behaviors. Augmenting the LLM with these explicit rationales (goals, constraints, trade-offs) enables interpretable, grounded reasoning and improves recommendation accuracy.}
  \label{fig:MARI}
\end{figure}

Integrating Large Language Models (LLMs) into recommender systems enables semantic user modeling by treating interaction sequences as contextual inputs. Current research predominantly serializes single-type interactions (e.g., ratings or purchases) into long textual prompts~\cite{bao2023tallrec, kim2024large, hou2024large,kim2025lost}. However, these sequences are costly to process and contain substantial redundancy and behavioral noise. While single-behavior datasets (e.g., Amazon~\cite{hou2024bridging}) have advanced the field, they overlook a critical real-world challenge: understanding the complex decision logic derived from heterogeneous, multi-behavior interactions (e.g., search, view, favorite, purchase) fundamental to e-commerce and streaming platforms.

Traditional multi-behavior models~\cite{gu2022self,cheng2023multi,lee2024mule,zhai2025combinatorial,zhang2025hierarchical} rely on implicit, opaque embeddings that lack the reasoning and interpretability strengths of LLMs. Consequently, the powerful inferential capabilities of LLMs remain underutilized for reconstructing user decision-making processes. Existing LLM-based recommenders either restrict themselves to single-behavior data or flatten multi-behavior logs into homogeneous sequences, losing the distinct semantic roles and hierarchical structures of different actions. Thus, the LLM revolution in recommendation has largely overlooked the rich decision signals woven across diverse behavioral contexts.

To bridge this gap, we propose an LLM-powered framework that explicitly reconstructs and leverages users' historical decision logic through a Decision Memory Bank (DMB), an interpretable repository of structured decision rationales. Our work is motivated by a key observation from industrial data: users' final decisions, especially when choosing between highly similar items (i.e., difficult-choice scenarios at the SKU level), reflect consistent patterns encoding their underlying criteria (e.g., prioritizing ``organic'' over ``budget'').
Because these latent patterns are distributed across multi-behavior trajectories, directly feeding raw sequences to LLMs is inefficient; it introduces noise and forces on-the-fly decision reconstruction under tight context constraints, a limitation visually demonstrated in the top workflow of Figure~\ref{fig:MARI}. In contrast, our two-stage approach (Figure~\ref{fig:MARI}, bottom) constructs the DMB, where each entry is a Structured Decision Memory (SDM), a concise textual record of a user's goals, constraints, and trade-offs. We devise Post-Hoc Decision Distillation (PHDD), which uses an LLM as a retrospective analyst to generate SDMs by examining final outcomes within heterogeneous behavioral contexts. This transforms noisy logs into explicit decision evidence. Second, we introduce MARI (Memory-Augmented Recommendation with Interpretability), a plug-and-play framework that retrieves relevant SDMs to ground LLM reasoning in explicit historical logic. This design cleanly separates offline knowledge construction from online recommendation, enabling robust, interpretable inference without significant latency overhead.

Our contributions are threefold:
\begin{itemize}
\item \textbf{Fine-Grained SKU-Level Recommendation:} We reformulate LLM-based recommendation as a SKU-level task, leveraging heterogeneous behaviors to reason over near-identical items. This effectively addresses difficult-choice scenarios with nuanced attribute differences, which are typically overlooked by title-level approaches.
\item \textbf{Decision Memory Bank via Retrospective Distillation:} We introduce the DMB, a compact archive of human-readable Structured Decision Memories (SDMs). Distilled via our Post-Hoc Decision Distillation (PHDD) method, each SDM encodes a user’s goals, constraints, and trade-offs from heterogeneous trajectories.
\item \textbf{Effective Memory-Augmented Recommendation Architecture:} We design a modular framework integrating the DMB via Retrieval-Augmented Generation (RAG) to enable interpretable reasoning grounded in historical logic. MARI achieves significant performance gains over state-of-the-art baselines in offline evaluations on a real-world e-commerce platform, demonstrating practical efficacy and scalability.
\end{itemize}

\section{Related Work}
\label{sec:related}
\subsection{LLM-based Recommendation}
\label{sec:related:llm4rec}
Large Language Models (LLMs) have enabled semantic user modeling in recommendation, moving beyond ID-based collaborative filtering~\cite{mnih2007probabilistic,chaney2015probabilistic, he2017neural,kang2018self,sun2019bert4rec,kim2023task}. Existing approaches fall into three categories: (1) feature enhancement, where LLMs enrich embeddings with textual semantics~\cite{ren2024representation, zhang2024notellm}; (2) end-to-end recommendation, where LLMs are prompted or tuned to predict preferences directly~\cite{bao2023tallrec, kim2024large, hou2024large,kim2025lost}; and (3) behavior compression and summary, where lengthy user histories are compressed into compact representations or summarized into semantic profiles to address context-length limitations~\cite{ning2025user, zheng2024harnessing, kim2408review}.
However, these methods either perform preference inference and summarization based on a single type of user behavior, or incorporate multiple behaviors only as semantically impoverished IDs, failing to account for their distinct roles in the decision-making process. A-LLMRec~\cite{hou2024large} and LLM-SRec~\cite{kim2025lost} align collaborative filtering signals with the LLM’s token space to inject collaborative awareness but operate on a single type of interaction, thus lacking the capacity to model cross-behavior decision dynamics; while User-LLM~\cite{ning2025user} compresses multiple behaviors into fixed-size vectors, obscuring transitions between action types. While converting raw behavior streams into semantic summaries effectively mitigates context-length and sparsity issues~\cite{zheng2024harnessing,kim2408review}, these remain shallow abstractions that reflect what users did, not why they decided. 
More fundamentally, static global summaries aggregate cross-category preferences, losing the context-dependent trade-offs essential for SKU-level difficult choices. We instead conditionally retrieve structured decision rationales to preserve this specificity without aggregation loss.

\subsection{Multi-Behavior Recommendation}
\label{sec:related:multibehavior}
Multi-behavior recommendation leverages heterogeneous interactions to mitigate data sparsity. While early work extends matrix factorization~\cite{zhao2015improving} or designs behavior-aware sampling~\cite{ding2018improving,guo2017resolving}, recent deep models use attention~\cite{guo2019buying,xia2020multiplex} or graph convolutions to fuse multi-behavior signals~\cite{gu2022self,cheng2023multi,lee2024mule,zhai2025combinatorial,zhang2025hierarchical}. Yet all produce opaque embeddings that capture statistical correlations without revealing the decision intent behind behavior transitions. Moreover, they ignore rich item textual attributes—such as price, brand, and product descriptions—that could ground user choices in semantic evidence.
Recent LLM-based approaches often concatenate heterogeneous actions into flat sequences~\cite{geng2024breaking}, failing to capture the decision-level meanings that different behavior types convey in the user’s choice process.
Critically, both lines predict next actions rather than modeling the decision logic behind behavior transitions—they capture statistical patterns (e.g., “view X → purchase Y”) but miss the underlying goals (e.g., seeking premium brands) or constraints (e.g., limited budget). We address this by using LLMs to retrospectively distill explicit decision rationales from trajectories, enabling recommendations grounded in interpretable evidence.

\subsection{Retrieval-Augmented Recommendation}
\label{sec:related:rag}
Retrieval-Augmented Generation (RAG)~\cite{lewis2020retrieval} reduces hallucination by grounding LLMs in external knowledge. In recommendation, methods like CoRAL~\cite{wu2024coral} and RETURN~\cite{ning2025retrieval} retrieve collaborative signals (e.g., co-purchases) to support long-tail reasoning. However, they retrieve only atomic interactions (“what users did”) without capturing the underlying decision logic (“why they decided”).
Unlike these approaches—which aim to improve item prediction by retrieving behavioral or item-level evidence—our framework targets the reconstruction of interpretable decision rationales from multi-behavior trajectories.
We argue that effective multi-behavior reasoning requires retrieving decision patterns, not just behavior logs. Our Decision Memory Bank (DMB) stores Structured Decision Memories (SDMs)—concise records of context, rationale, and outcome—so that RAG can retrieve historically grounded, trade-off-aware decision evidence instead of surface-level co-occurrences.

\section{Preliminary}
Let $\mathcal{U}$ and $\mathcal{I}$ denote the sets of users and items, respectively. For a user $u \in \mathcal{U}$, we observe a historical heterogeneous interaction sequence:
\begin{equation}
\mathcal{H}_u = \left\{ (i_t, a_t, \tau_t) \right\}_{t=1}^{T_u},
\end{equation}
where $i_t \in \mathcal{I}$ is the interacted item (associated with rich attributes and metadata), $a_t \in \mathcal{A}$ is the associated behavior type (e.g., $\mathcal{A} = \{\text{view}, \text{search}, \text{favorite}, \text{purchase}\}$) and $\tau_t$ is the timestamp. Interactions are chronologically ordered ($\tau_1 < \tau_2 < \cdots < \tau_{T_u}$). Notably, items in $\mathcal{I}$ are defined at the SKU level—distinct attribute combinations (e.g., color or size) yield distinct items, even with identical titles. An example is given in Appendix~\ref{app:item}.

In real-world recommendation scenarios, users typically exhibit multiple types of behavior before making a final decision (e.g., purchase). Critically, these heterogeneous behaviors encode complementary signals about the user's underlying intent—such as exploration (search/view), preference expression (favorite), and commitment (purchase). However, directly modeling $\mathcal{H}_u$ as a flat token sequence overlooks the semantic hierarchy among action types and fails to reconstruct the high-level decision logic that governs user choices.
To systematically evaluate a model's capacity to overcome this limitation and capture such latent decision logic, we formalize two complementary prediction tasks that reflect different levels of recommendation difficulty and intent granularity.

\subsection{Task 1: Next-Item Prediction (NIP)}
Given the full historical sequence $\mathcal{H}_u$ and a recent short-term context $\mathcal{C}_u \subseteq \mathcal{H}_u$ (e.g., the last $k$ interactions), the goal is to predict the next item the user will purchase. In practice, a purchased item always appears earlier in $\mathcal{C}_u$ through non-purchase interactions (e.g., view or search). Thus, NIP requires the model to identify the specific item within $\mathcal{C}_u$ that will eventually be purchased. This task assesses the model's capacity to identify the next purchase from the full item set using the user’s heterogeneous interaction history.

\subsection{Task 2: Difficult Choice Prediction (DCP)}
This task simulates realistic “difficult-choice” scenarios where users compare highly similar alternatives (e.g., products with near-identical specifications but differing in price, brand, or minor features). Formally, given $\mathcal{H}_u$, a recent context $\mathcal{C}_u$ , and a candidate set $\mathcal{X} \subset \mathcal{I}$ of $m$ items with high pairwise similarity, the goal is to predict which item in $\mathcal{X}$ the user will purchase next. Unlike NIP, DCP isolates the ability to resolve fine-grained user preferences by restricting candidate items to highly substitutable sets, thereby directly testing a model's capacity to infer subtle trade-offs—such as “willingness to pay more for organic certification”—from past decisions.

\subsection{Key Challenge and Our Approach}
\begin{figure}[htbp]
\centering
\includegraphics[width=0.8\linewidth]{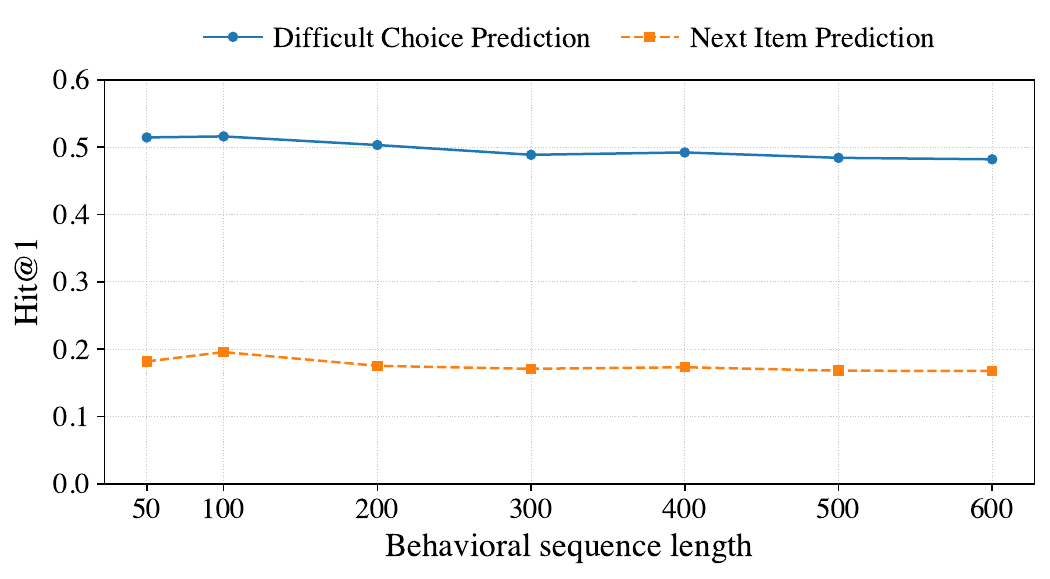}
\caption{Effect of input sequence length on Hit@1 for NIP and DCP tasks (using DeepSeek-R1).}
\label{fig:context_length}
\end{figure}

Both NIP and DCP require distilling latent decision logic—such as goals, constraints, and attribute-level trade-offs—from the noisy heterogeneous sequence $\mathcal{H}_u$, rather than relying on surface-level patterns. However, as shown in Figure~\ref{fig:context_length}, our experiments with DeepSeek-R1~\cite{deepseekai2025deepseekr1incentivizingreasoningcapability} show that simply extending the input context yields diminishing returns: performance peaks at around 100 historical interactions and degrades with longer sequences, suggesting that raw behavior logs overwhelm LLMs with noise rather than useful signals.

To address this, we propose a personalized Decision Memory Bank (DMB) that stores Structured Decision Memories (SDMs), generated via Post-Hoc Decision Distillation (PHDD) as concise, interpretable summaries of past decisions. Building on DMB, we design MARI (Memory-Augmented Recommendation with Interpretability), which retrieves relevant SDMs during inference to ground the LLM’s reasoning about $\mathcal{C}_u$ in explicit decision logic. Formal details follow.

\section{Methodology}
\begin{figure*}[htbp]
  \centering
  \includegraphics[width=\textwidth]{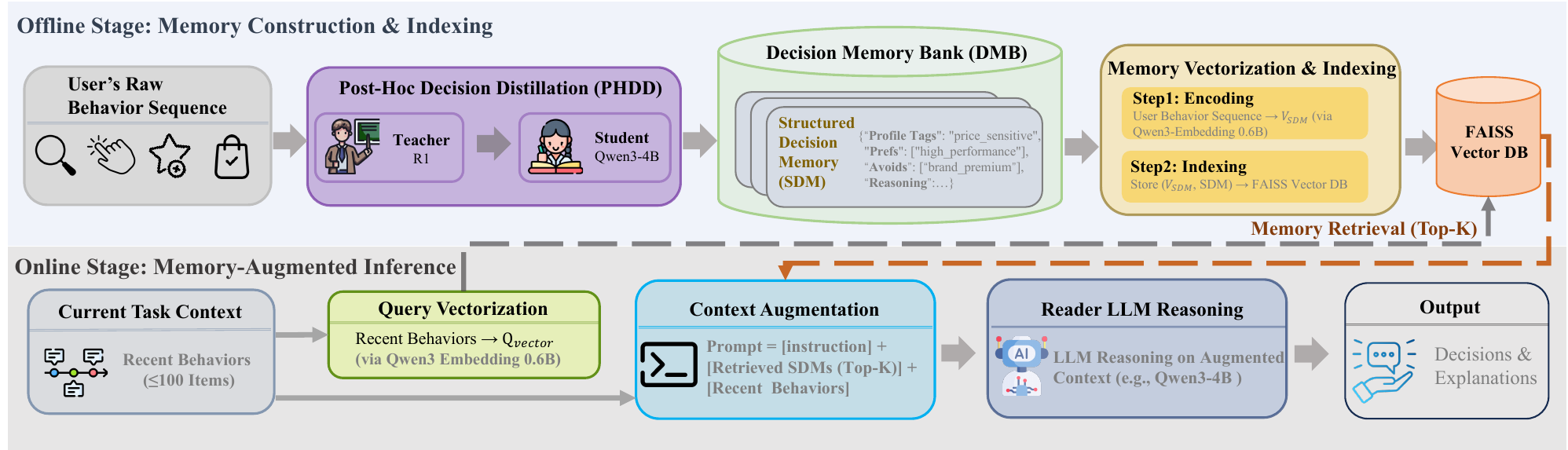}
  \caption{Algorithmic framework of MARI.
The offline stage constructs DMB via PHDD; during online inference, MARI retrieves relevant SDMs and fuses them to prompt the LLM for interpretable recommendation.}
  \label{fig:MARI_FRAMEWORK}
\end{figure*}
We present MARI, a two-stage framework that decouples decision knowledge construction from on-the-fly recommendation reasoning. As illustrated in Figure~\ref{fig:MARI_FRAMEWORK}, MARI first builds a personalized DMB for each user via Post-Hoc Decision Distillation (PHDD), then leverages retrieval-augmented generation to inject relevant decision memories into an LLM-based recommender during inference. Below we detail the DMB construction pipeline and the MARI inference architecture for both NIP and DCP tasks.

\subsection{Decision Memory Bank (DMB)}
\label{subsec:DMB}
\subsubsection{Category-Aware}
While some decision tendencies (e.g., cost-effectiveness) generalize across categories, purchase logic is largely category-specific: “Organic certification” matters for baby food but not electronics, whereas “512GB storage” drives electronics choices but not food purchases. Flattening cross-category interactions dilutes these discriminative, category-conditioned rules. To preserve fidelity, we build the DMB in a category-aware manner, with each entry encoding the structured rationale behind a user’s purchase in category $c \in \mathcal{C}$.

\subsubsection{Formal definition}
For user $u$, the Decision Memory Bank is defined as a set of structured records:
\begin{equation}
\text{DMB}_u = \bigcup_{c \in \mathcal{C}} \left\{ \left( \mathbf{e}_{u,c}^{(j)}, \mathcal{M}_{u,c}^{(j)}, \tau_{u,c}^{(j)} \right) \right\}_{j=1}^{J_{u,c}}
\end{equation}
where $J_{u,c}$ is the number of purchases user $u$ made in category $c$; $\tau_{u,c}^{(j)}$ is the timestamp of the $j$-th purchase; $\mathbf{e}_{u,c}^{(j)}\in\mathbb{R}^d$ is the embedding vector of the pre-purchase heterogeneous behavior sequence leading to this purchase (used for similarity-based retrieval); and $\mathcal{M}_{u,c}^{(j)}$ is the structured decision memory (SDM) —a structured quintuple capturing five complementary facets of the user's decision logic:
\begin{equation}
\mathcal{M}_{u,c}^{(j)} = \big( \mathcal{R}_{u,c}^{(j)}, \mathcal{T}_{u,c}^{(j)}, \mathcal{E}_{u,c}^{(j)}, \mathcal{P}_{u,c}^{(j)}, \mathcal{A}_{u,c}^{(j)} \big).
\end{equation}
Each component is formally defined as follows:
\begin{itemize}[leftmargin=*]
\item Decision Intent Reasoning $\mathcal{R}_{u,c}^{(j)}$: a natural language explanation of the purchase of item $i^*$, grounded in the user’s historical heterogeneous behaviors.
\item Decision Profile Tags $\mathcal{T}_{u,c}^{(j)}$: a set of categorical labels characterizing the user's decision style in this context, drawn from a predefined taxonomy $V_{\mathrm{tag}}$. Examples of user decision profile tags include brand-sensitive, price-sensitive, and certification-driven preferences.
\item Decision Efficiency $\mathcal{E}_{u,c}^{(j)}$: a discrete decision tempo type reflecting the user’s decision-making speed, inferred from behavioral signals such as session duration, number of item comparisons, and backtracking frequency. This yields four categories of decision efficiency: flash, rapid, considered, and prolonged.
\item Decision Preference Factors $\mathcal{P}_{u,c}^{(j)}$: a set of key-value pairs $\{(k_p,\allowbreak v_p)\}$ where $k_p$ denotes a positively weighted decision attribute (e.g., brand, price, style) and $v_p$ specifies its concrete instantiation in the chosen item. Example: {(brand, Nike), (price, 500--800), (style, minimalist)}.
\item Decision Avoidance Factors $\mathcal{A}_{u,c}^{(j)}$: a set of key-value pairs $\{(k_a,\allowbreak v_a)\}$ capturing attributes explicitly rejected during exploration. Here $k_a$ denotes a decision dimension (same category as $k_p$), and $v_a$ specifies the avoided value. Example: {(material, polyester), (price, $>$1200), (color, bright-red)}.
\end{itemize}
Both the profile tags $\mathcal{T}_{u,c}^{(j)}$ and the keys in $\mathcal{P}_{u,c}^{(j)}$ and $\mathcal{A}_{u,c}^{(j)}$ are from closed vocabularies, derived via iterative LLM-based clustering on large-scale e-commerce data. The LLM also generates evidence-backed rationales for each factor to ensure plausibility. An example of an SDM is provided in Appendix~\ref{app:SDM}.

\subsection{ Post-Hoc Decision Distillation (PHDD)}
\label{subsec:PHDD}
To enable efficient inference, we design a two-stage distillation pipeline that yields a lightweight student model for generating user SDMs: (1) a teacher LLM generates structured memories and Chain-of-Thought (CoT)~\cite{wei2022chain}; (2) a small student model is trained via CoT distillation~\cite{magister2023teaching} to mimic this capability. 

\subsubsection{Teacher Model: Category-Conditioned Memory Generation}
To construct high-quality CoT training data for PHDD, we leverage the powerful reasoning capability of DeepSeek-R1~\cite{deepseekai2025deepseekr1incentivizingreasoningcapability} to generate grounded, interpretable decision rationales from real user interaction logs. Specifically, for each purchase event $(i^*, c, \tau^*)$ by user $u$ in category $c$, we extract the category-filtered pre-purchase sequence:
\begin{equation}
\mathcal{H}_{u,c}^{\text{pre}} = \left\{ (i_t, a_t, \tau_t) \mid \tau_t < \tau^*, \text{category}(i_t) = c \right\}.
\end{equation}

We employ a teacher LLM $\phi^{\text{PHDD}}_{\text{tea}}$, prompted with a carefully designed template to enforce a consistent output format, generating structured decision memories $\mathcal{M}_{u,c}^{(j)}$. These outputs are subsequently verified to eliminate hallucinations and ensure high-quality reasoning trajectories. Critically, to enhance rationale accuracy, we provide the teacher with post-purchase reviews as privileged calibration references exclusively during dataset construction. Guided by the prompt, the teacher deduces decision logic solely from pre-purchase behaviors, utilizing the reviews merely to validate the inferred rationale without directly leaking their text. These signals are excluded during student training and inference, ensuring realistic deployment conditions.

\subsubsection{Student Model Training via CoT Distillation}
To enable scalable DMB construction in real-world production scenarios, we distill the teacher’s capability into a compact student LLM $\phi^{\text{PHDD}}_{\text{stu}}$ (Qwen3-4B~\cite{qwen3}) using Chain-of-Thought distillation:
\begin{itemize}[leftmargin=*]
\item Dataset creation: For each purchase event, we collect the teacher’s CoT reasoning trace (intermediate rationale steps) and final structured memory $\mathcal{M}_{u,c}^{(j)}$ as supervision signals.
\item Input representation: The student receives only $\mathcal{H}_{u,c}^{\text{pre}}$ and category context $c$.
\item Training objective: Minimize the cross-entropy loss between student-generated tokens and the teacher’s CoT trace + final memory:
\begin{equation}
\mathcal{L}_{\text{PHDD}} = -\sum_{l=1}^{L} \log P_{\phi^{PHDD}_{\text{stu}}}\left( y_l^{\text{tea}} \mid \mathcal{H}_{u,c}^{\text{pre}}, c, y_{<l}^{\text{tea}} \right)
\end{equation}
where $y_{1:L}^\text{tea}$ is the concatenated CoT trace and SDM text.
\end{itemize}

After training, $\phi^{\text{PHDD}}_{{\text{stu}}}$ generates SDMs $\mathcal{M}_{u,c}^{(j)}$ for all historical purchases. For each SDM, we compute its retrieval embedding $\mathbf{e}_{u,c}^{(j)}$ by encoding the corresponding pre-purchase sequence $\mathcal{H}_{u,c}^{\text{pre}}$ with a lightweight, frozen text encoder (e.g., Qwen3-Embedding-0.6B~\cite{qwen3embedding}). The resulting $\text{DMB}_u$ is stored as a category-indexed vector database supporting efficient similarity search.

\subsection{Memory-Augmented Reasoning Framework}
\label{subsec:RAG}
The MARI inference pipeline consists of two core stages: (1) context processing and adaptive memory retrieval, and (2) prompt construction and memory-augmented LLM reasoning.

\subsubsection{Context Processing and Adaptive Memory Retrieval}
Given a user's recent interaction context $\mathcal{C}_u = \{(i_t, a_t, \tau_t)\}_{t=T_u-k+1}^{T_u}$, we first encode it into a dense vector $\mathbf{e}_{\text{ctx}}$ using the frozen text encoder (same as in DMB construction). The retrieval strategy then adapts to the prediction task.

For Next-Item Prediction (NIP): The target category is unknown a priori. We perform cross-category similarity retrieval to find SDM with behaviorally similar exploration patterns:

\begin{equation}
\label{eq:recall}
\mathcal{S}_u^{\text{NIP}} = 
\left\{ 
  \mathcal{M}_{u,c}^{(j)} 
  \,\middle|\, 
  (c,j) \in 
  \underset{
    \substack{
      c \in \mathcal{C}, j \in [1, J_{u,c}]
    }
  }{\operatorname{TopK}} 
  \left( 
    \text{sim}\left( 
      \mathbf{e}_{\text{ctx}},\, 
      \mathbf{e}_{u,c}^{(j)} 
    \right) 
  \right) 
\right\}
\end{equation}
where $\text{sim}(\mathbf{a},\mathbf{b}) = \frac{\mathbf{a}^\top \mathbf{b}}{\|\mathbf{a}\|\|\mathbf{b}\|}$.

For Difficult Choice Prediction (DCP): The target category $c_{\text{target}}$ is known from session context. We perform category-constrained temporal retrieval:
\begin{equation}
\mathcal{S}_u^{\text{DCP}} = \left\{ \mathcal{M}_{u,c_{\text{target}}}^{(j)} \mid j \in \text{TopK}_{\text{time}} \left( \{ \tau_{u,c_{\text{target}}}^{(j)} \} \right) \right\}.
\end{equation}
Prioritizing recency over behavioral similarity is critical here: fine-grained preference trade-offs evolve rapidly within categories and are best captured by recent decisions. If the user has no SDM records in the target category $c_{\text{target}}$, we fall back to decision memory retrieval as defined in Equation~\ref{eq:recall}.

\subsubsection{Prompt Construction and Memory-Augmented Reasoning}
\label{sub:prompt}
We integrate retrieved memories $\mathcal{S}_u$ with the current context $\mathcal{C}_u$ to construct task-specific prompts for $\phi^{\text{rec}}$, an LLM-based recommender.
\begin{itemize}[leftmargin=*]
\item For NIP, where the target category is unknown a priori, the prompt instructs the LLM to predict the next purchased item directly from the user’s recent interaction sequence, augmented with SDMs retrieved from the DMB based on the current behavioral context.

\item For DCP, where the candidate set within a known category is provided, the prompt instructs the LLM to predict the next purchased item directly from the candidate set, augmented with SDMs retrieved from the DMB preferentially based on the target category.
\end{itemize}

Both designs enforce structured reasoning grounded in explicit decision evidence, enabling $\phi^{\text{rec}}$ to move beyond surface-level sequential patterns—enhancing robustness under data sparsity while preserving interpretability through traceable memory references.
The prompt template is provided in Appendix~\ref{app:prompt}.

\subsection{Deployment Strategies}
\label{subsec:DS}
MARI supports two deployment paradigms that balance training overhead, interpretability, and integration flexibility.

\subsubsection{Plug-and-Play Augmentation Mode}
MARI functions as a lightweight, training-free augmentation module for existing LLM-based recommenders. Given context $\mathcal{C}_u$, it retrieves relevant memories $\mathcal{S}_u$ and constructs the task-appropriate prompt for any off-the-shelf recommendation LLM $\phi^{\text{rec}}$. This training-free compatibility enables:
\begin{itemize}[leftmargin=*]
    \item Seamless integration into production pipelines.
    \item No additional training costs or architectural modifications.
    \item Preservation of existing model investments while enhancing decision transparency.
\end{itemize}

\subsubsection{CoT-Distilled Mode}
For maximum performance and full interpretability, we distill a specialized student model $\phi_{\text{stu}}^{\text{rec}}$ that generates both predictions and justifications. Training follows the MARI inference pipeline:
     (1) For each purchase $(i^*, \tau^*)$, extract pre-purchase context $\mathcal{C}_u^{\text{pre}}$;
  (2) Retrieve relevant memories $\mathcal{S}_u^{\text{pre}}$ using task-appropriate strategy;
    (3) Prompt the teacher LLM to generate CoT reasoning from 
$(\mathcal{C}_u^{\text{pre}}, \mathcal{S}_u^{\text{pre}})$, concluding with $i^*$;
     (4) Train student to autoregressively generate this CoT+item sequence label $y^{rec}$:
\begin{equation}
\mathcal{L}_{\text{rec}} = -\sum_{l=1}^{L} \log P_{\phi^{rec}_{\text{stu}}}\left( y^{rec}_l \mid \mathcal{C}_u^{\text{pre}} \oplus \mathcal{S}_u^{\text{pre}}, y^{rec}_{<l} \right).
\end{equation}


At inference, $\phi_{\text{stu}}^{\text{rec}}$ outputs human-interpretable rationales alongside predictions. This establishes a new paradigm for transparent recommendation where every prediction is grounded in explicit, retrievable decision evidence.

Both deployment modes share the same DMB infrastructure, allowing practitioners to choose based on their system constraints and interpretability requirements. This design enables MARI to scale efficiently to large-scale systems while preserving detailed decision semantics critical for difficult-choice recommendations.

\section{Experiments Setup}
\subsection{Datasets}
Existing public multi-behavior datasets lack the rich textual item metadata (e.g., titles, attributes) essential for LLM-based reasoning, as they mostly contain anonymized IDs. To address this, we construct a new large-scale dataset from six months of real-world user logs on a major e-commerce platform, capturing heterogeneous behaviors—search, view, favorite, and purchase—along with comprehensive item metadata. We further split it into two task-specific datasets: the Post-Hoc Decision Distillation (PHDD) Dataset and the Multi-Behavior Recommendation (MBR) Dataset.

\subsubsection{PHDD Dataset}
The PHDD Dataset captures fine-grained, user-level decision traces in their original form, preserving all observed interactions without behavioral filtering or deduplication. It is partitioned into 200k, 10k, and 10k samples for training, validation, and testing, respectively.

\subsubsection{MBR Dataset}
We construct this dataset to evaluate tasks NIP and DCP, following the previous methods~\cite{gao2019learning} with two preprocessing steps to ensure data quality: (1) we merge duplicated user-item-behavior interactions by keeping the earliest record; (2) we apply a k-core filtering (k=10) to retain users and items with at least 10 interactions, mitigating data sparsity. Each user’s filtered chronological sequence is then split via leave-one-out by timestamp: the last purchase for testing, the second-last for validation, and earlier interactions for training.

To evaluate the cross-category generalization capability of our MARI framework under the Next Item Prediction (NIP) setting, we further partition the instances into two disjoint subsets: (1) \textbf{NIP-in-category (NIP-in)}, comprising users who have previously purchased at least one item in the same category as the target item; and (2) \textbf{NIP-out-of-category (NIP-out)}, where the target purchase represents the user's first interaction in that category. This partition enables us to assess whether our method can effectively leverage decision memory to support recommendations in both familiar and novel category contexts.

For the DCP task, where the target item category is assumed to be known, we construct the candidate set by using an LLM (Qwen3-Max) to select 2–5 items within the same category that the user showed notable hesitation or comparison behavior toward in their recent interaction history, followed by manual validation of the selected candidates. An example of a candidate set is provided in Appendix~\ref{app:candi}.

The resulting MBR dataset comprises three distinct tasks (NIP-in, NIP-out, and DCP) with non-overlapping users across tasks, yielding a total of 150,000 unique users. To ensure fair comparison, each task is identically sampled into 50k training, 10k validation, and 10k testing instances.

\subsection{Metrics}
\subsubsection{PHDD Task}
For the PHDD task, Decision Profile Tags, Decision Efficiency, and the keys of Decision Preference/Avoidance Factors are closed-set. We evaluate them using precision, defined as the fraction of correctly predicted items among all predictions.
In contrast, Decision Intent Reasoning and the values of Preference/Avoidance Factors are open-ended. We assess them via an LLM-as-judge approach, measuring the following dimensions on a scale of 0–5:
(1) structural correctness (adherence to output format),
(2) reasoning quality (logical coherence of the decision inference process), and
(3) value abstraction (appropriateness and generalizability of extracted factor values).
Specifically, we employ DeepSeek-R1 as the judge model and provide it with a human-verified reference solution as the scoring benchmark to ensure evaluation reliability.
\subsubsection{MBR Task}
For the MBR task, we adopt a next-item prediction setting where the user's subsequent actual purchase serves as the ground-truth target. Model performance is evaluated using a
widely used metric Hit Ratio@1 (Hit@1).

\subsection{Experimental Settings}
\subsubsection{Baseline}
We establish a comprehensive set of baseline models for our two core tasks.

For the PHDD task, which focuses on generating a user's decision memory, we adopt the off-the-shelf Qwen3~\cite{qwen3} series of large language models as our primary baselines. These models are used without any additional fine-tuning.

For the MBR task, we benchmark against two baseline categories. First, we include ID-based sequential recommenders: $\text{SASRec}_{MB}$~\cite{yang2022multi}, SASRec's~\cite{kang2018self} multi-behavior extension that incorporates different interaction types; and three recent state-of-the-art ID-based multi-behavior recommendation methods, namely MULE~\cite{lee2024mule}, COPF~\cite{zhai2025combinatorial}, and HGIB~\cite{zhang2025hierarchical}.
Second, to fully leverage the rich textual semantics of items, we integrate several powerful open-source LLMs into the recommendation pipeline. Specifically, we employ Qwen3, DeepSeek-R1 (R1)~\cite{deepseekai2025deepseekr1incentivizingreasoningcapability}, and GPT-OSS~\cite{openai2025gptoss120bgptoss20bmodel}, which process item metadata such as titles, prices, and attributes using their native language understanding capabilities.
Similar to the PHDD task, these LLMs are also used in their original, unmodified form without any further training on recommendation data. To ensure fair comparison, all baselines predict the target item from a unified candidate space: the recent historical sequence (for NIP) or the provided candidate set (for DCP).

\subsubsection{Implementation Details}
For the PHDD task, we set the maximum length of the input behavior sequence $\mathcal{H}_{u,c}$ to 200, considering only interactions that occurred within a two-week window prior to the target decision.
To support retrieval in the DMB, we employ the Qwen3-Embedding-0.6B model to extract dense representations of behavior sequences. 
The PHDD Model $\phi^{\text{PHDD}}_{\text{stu}}$ is trained via full-parameter fine-tuning on the PHDD dataset, based on the Qwen3-4B.

For the MBR task, we set the maximum input sequence length $\mathcal{C}_u$ to 100, as our preliminary experiments (Figure~\ref{fig:context_length}) show that model performance peaks at this length. 
During memory retrieval, we perform a top-$K$ search with $K=3$, retrieving the three most relevant decision memories from the DMB. 
The MBR model $\phi^{\text{rec}}_{\text{stu}}$ is fully fine-tuned from Qwen3-4B on the constructed MBR dataset. We train the model on a mixture of the three tasks: NIP-in-category, NIP-out-of-category, and DCP.

\section{Experimental Results}
\subsection{Performance Comparison}
We perform comprehensive performance comparisons across two distinct tasks: the PHDD task, designed to generate users' structured decision memories, and the Multi-Behavior Recommendation (MBR) task.

\subsubsection{PHDD}
Results on the PHDD dataset (Table~\ref{table:pdd}) show that our method enables the compact Qwen3-4B to match the much larger R1 model in precision metrics, while surpassing it in Reasoning Quality and Value Abstraction under LLM-as-Judge evaluation. This confirms that PHDD effectively distills the ability to understand user decision intent into a smaller model, ensuring the generation of high-quality SDMs while achieving competitive performance at a fraction of the inference cost.

\begin{table}[htbp] 
\centering
\caption{Quality of Structured Decision Memory (SDM) generation on the PHDD dataset. \textbf{Models}: Q4B (Qwen3-4B), Q8B (Qwen3-8B), Q14B, Q32B. \textbf{Bold}: Best, \underline{Underlined}: Second-best.}
\label{tab:phdd_single}
\footnotesize
\setlength{\tabcolsep}{3.5pt} 
\begin{tabular}{@{} l *{6}{c} @{}}
\toprule
\textbf{Metric} & \textbf{Q4B} & \textbf{Q8B} & \textbf{Q14B} & \textbf{Q32B} & \textbf{R1} & \textbf{Q4B-PHDD(Ours)} \\
\midrule
\multicolumn{7}{@{}l}{\textit{Precision ($\uparrow$)}} \\
Factor Key      & 0.577 & 0.654 & 0.636 & 0.631 & \textbf{0.779} & \underline{0.769} \\
Decision Prof.  & 0.467 & 0.470 & 0.525 & 0.556 & \textbf{0.725} & \underline{0.713} \\
Decision Effic. & 0.483 & 0.473 & 0.529 & 0.513 & \textbf{0.707} & \underline{0.693} \\
\midrule
\multicolumn{7}{@{}l}{\textit{LLM-as-Judge ($\uparrow$)}} \\
Struct. Corr.   & 3.56  & 3.31  & 3.58  & 3.71  & \textbf{4.13}  & \underline{4.12} \\
Reason. Qual.   & 3.51  & 3.40  & 3.69  & 3.75  & \underline{3.89} & \textbf{3.93} \\
Value Abst.     & 3.49  & 3.50  & 3.66  & 3.81  & \underline{4.21} & \textbf{4.22} \\
\bottomrule
\end{tabular}
\label{table:pdd}
\end{table}

\subsubsection{MBR}

Table \ref{table:main} details the Hit@1 performance, revealing three core findings:
\textbf{(1) Plug-and-play SDMs generalize across diverse model scales.} By simply injecting SDMs into off-the-shelf LLMs without weight updates, MARI$^\dagger$ broadly enhances performance across diverse parameter sizes (from 4B to 671B). Crucially, when applied to sufficiently large models (e.g., R1, Qwen3-235B-A22B-Think), this plug-and-play approach successfully surpasses strong fully-trained ID-based baselines (e.g., HGIB, MULE), explicitly confirming the generality and potency of our framework.
\textbf{(2) SDMs significantly amplify reasoning on complex tasks.} The structured priors provide vital context that medium-to-large models ($\ge$14B) can effectively leverage, yielding massive gains on the challenging NIP-out task (e.g., +165\% for GPT-OSS-20B, +128\% for Qwen3-32B). However, smaller models (4B/8B) degrade on NIP-out, suggesting that out-of-category reasoning inherently requires a minimum intrinsic capacity. Conversely, on the DCP task, MARI$^\dagger$ empowers all model scales to consistently outperform ID-based baselines, highlighting LLMs' inherent advantage in difficult choice prediction that demands semantic comparison.
\textbf{(3) SFT empowers compact models to achieve state-of-the-art results.} Fine-tuning the lightweight 4B model yields substantial improvements across all three tasks. Notably, MARI (4B) achieves 0.5561 on NIP-out and 0.8868 on DCP, significantly outperforming both fully-trained ID-based models and massive 671B untuned LLMs like R1. This demonstrates that specializing compact models to reason over SDMs offers a practical pathway to superior accuracy, mitigating reliance on massive architectures.

\begin{figure}[htbp]
\centering
\includegraphics[width=\linewidth]{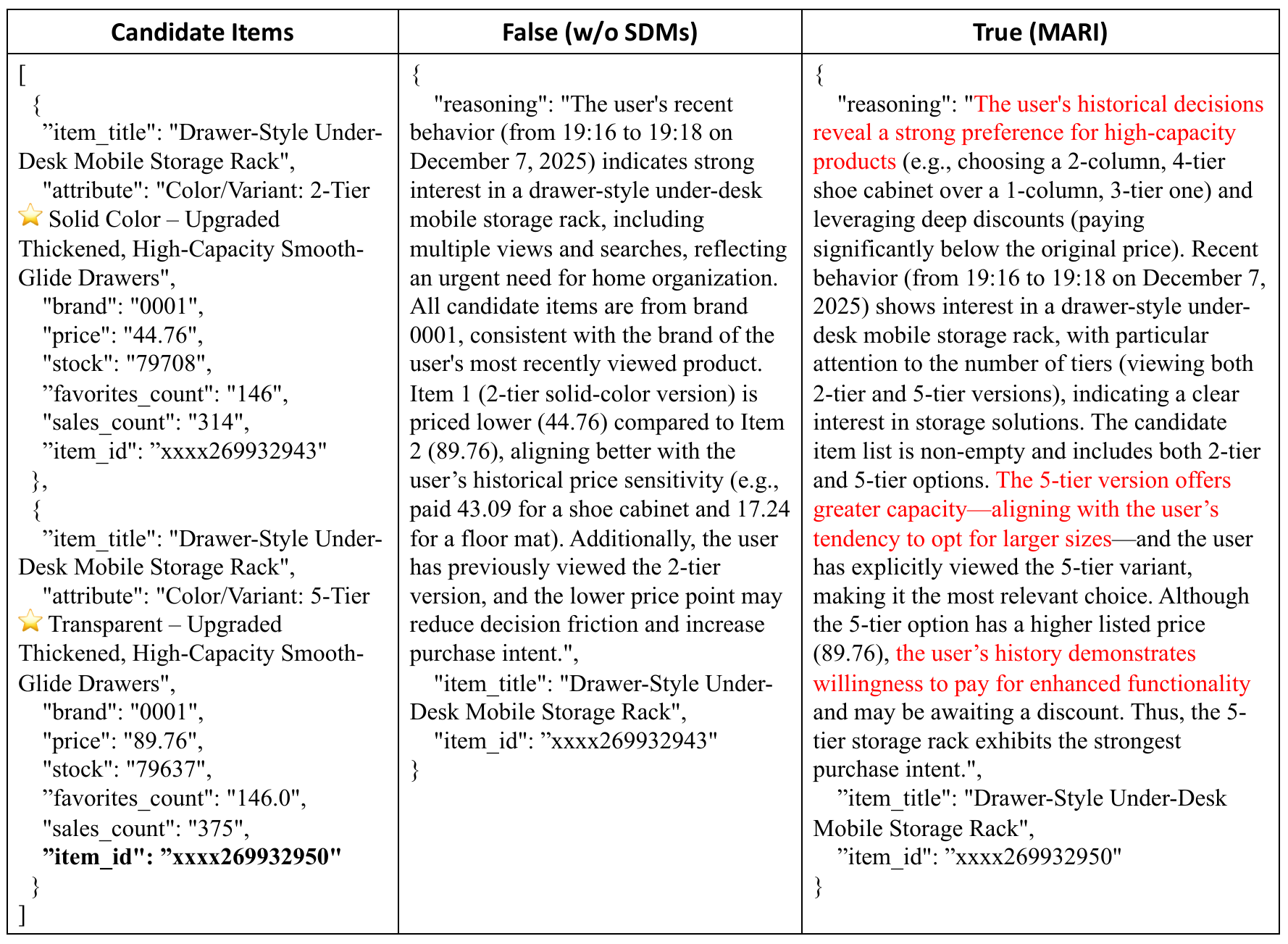}
\caption{The role of structured decision memory: comparing recommendation rationales with and without decision memory augmentation.}
\label{fig:result}
\end{figure}

Figure~\ref{fig:result} presents an example of MARI’s interpretable recommendations, demonstrating that the model effectively uses the information of the SDMs (marked in red) during reasoning to produce the correct product recommendations.

\begin{table*}[!t] 
\centering
\caption{Full comparison of Hit@1. MARI$^{\dagger}$ denotes an off-the-shelf open-source general-purpose model without any task-specific fine-tuning for recommendation. Base denotes vanilla LLMs and ID-based models. \textbf{Bold}: Best, \underline{Underlined}: Second-best.}
\label{tab:full_results} 
\footnotesize
\begin{tabular*}{\textwidth}{@{\extracolsep{\fill}} l l *{3}{c} @{\hspace{2em}} *{3}{c} @{} } 
\toprule
\multirow{2}{*}{\textbf{Category}} & \multirow{2}{*}{\textbf{Model}} & \multicolumn{3}{c}{\textbf{Base}} & \multicolumn{3}{c}{\textbf{MARI$^\dagger$ (Ours, w/o SFT)}} \\
\cmidrule{3-5} \cmidrule{6-8} 
 & & NIP-in & NIP-out & DCP & NIP-in & NIP-out & DCP \\
\midrule
\multirow{5}{*}{\makecell[l]{ID-based \\ (w/ train)}} 
 & SASRec$_{MB}$ & 0.1666 & 0.1697 & 0.4077 & - & - & - \\
 & MULE & 0.2363 & 0.1968 & 0.4206 & - & - & - \\
 & COPF & 0.2013 & 0.1795 & 0.4037 & - & - & - \\
 & HGIB & 0.2489 & 0.2251 & 0.4384 & - & - & - \\ 
\midrule
\multirow{8}{*}{\makecell[l]{LLMs \\ (w/o train)}} 
 & Qwen3-4B & 0.0751 & 0.0225 & 0.4617 & 0.1278 & 0.0157 & 0.4751 \\
 & Qwen3-8B & 0.1120 & 0.0687 & 0.4203 & 0.1580 & 0.0675 & 0.4505 \\
 & Qwen3-14B & 0.0999 & 0.0391 & 0.4514 & 0.1535 & 0.0676 & 0.4549 \\
 & GPT-OSS-20B & 0.1477 & 0.0934 & 0.4207 & 0.1963 & 0.2480 & 0.4533 \\
 & Qwen3-32B & 0.1136 & 0.0496 & 0.4479 & 0.1742 & 0.1135 & 0.4697 \\
 & Qwen3-235B-A22B-Ins & 0.1304 & 0.0798 & 0.5069 & 0.2224 & 0.2422 & 0.5251 \\
 & Qwen3-235B-A22B-Thk & 0.2068 & 0.2992 & 0.5010 & \underline{0.2687} & \underline{0.4189} & 0.5249 \\ 
 & R1 & 0.1957 & 0.2503 & 0.5158 & 0.2531 & 0.3641 & \underline{0.5336} \\ 
\midrule 
\textbf{MARI (Ours, w/ SFT)} & & - & - & - & \textbf{0.4869} & \textbf{0.5561} & \textbf{0.8868} \\ 
\bottomrule
\end{tabular*}
\label{table:main}
\end{table*}

\subsection{Ablation Study}
To validate the core design of MARI, we conduct a series of ablation studies comparing MARI-B, MARI$^{\dagger\star}$, MARI$^{\dagger}$, and SFT variants. Table~\ref{tab:ablation} details the configurations and results, revealing three core insights:

\textbf{(1) Superiority of structured memories over raw behaviors.} Directly injecting raw sequences aligned with SDMs (MARI-B) yields gains on simpler tasks (e.g., NIP-in) but degrades performance on complex tasks like NIP-out and DCP. This indicates that the verbosity of raw behaviors dilutes contextual information, distracting the LLM from critical decision factors required for deep reasoning. Conversely, using decision memories (MARI$^{\dagger\star}$ and MARI$^{\dagger}$) consistently outperforms MARI-B across all scales. This proves that enriching the input with explicit decision logic—rather than relying solely on implicit patterns in raw sequences—is crucial for guiding LLMs in challenging scenarios.

\textbf{(2) Critical impact of memory quality.} High-quality memories distilled via PHDD (MARI$^{\dagger}$) always beat undistilled, lower-quality ones (MARI$^{\dagger\star}$), particularly in NIP-out and DCP. This direct correlation between memory quality and performance validates the necessity of PHDD for filtering noise and abstracting clear user intent.

\textbf{(3) SDMs enhance supervised fine-tuning.} The baseline SFT and MARI share the same training objective, differing only in input augmentation. Incorporating retrieved SDMs during SFT yields clear gains over standard SFT without SDMs. This confirms a strong synergy between structured memory inputs and parameter updates for complex reasoning and generalization.

\begin{table}[htbp] 
\centering
\caption{Ablation study on Hit@1. \textit{MARI-B}: Uses raw behavior sequences aligned with SDMs. \textit{MARI$^{\dagger\star}$}: Uses undistilled SDMs generated by Qwen3-4B (w/o PHDD). \textit{MARI$^{\dagger}$}: Uses distilled SDMs generated by Qwen3-4B (w/ PHDD). \textit{SFT variants}: Compares standard SFT against SFT with SDMs. \textbf{Bold} indicates the best (MARI$^\dagger$) and overall (MARI) results.}
\label{tab:ablation}
\footnotesize
\setlength{\tabcolsep}{3.5pt} 
\begin{tabular}{@{} l *{5}{c} @{}}
\toprule
\textbf{Variant} & \textbf{Q4B} & \textbf{Q14B} & \textbf{Q32B} & \textbf{GPT-OSS-20B} & \textbf{R1} \\
\midrule
\multicolumn{6}{@{}l}{\textit{NIP-in}} \\
Base & 0.075 & 0.100 & 0.114 & 0.148 & 0.196 \\
MARI-B (Raw Behav.) & 0.105 & 0.125 & 0.127 & 0.132 & 0.247 \\
MARI$^{\dagger\star}$ (Undist. SDM) & 0.110 & 0.134 & 0.157 & 0.170 & 0.249 \\
\textbf{MARI$^\dagger$} (Dist. SDM) & \textbf{0.128} & \textbf{0.154} & \textbf{0.174} & \textbf{0.196} & \textbf{0.253} \\
SFT w/o SDM & 0.458 & - & - & - & - \\
\textbf{MARI} (SFT + SDM) & \textbf{0.487} & - & - & - & - \\
\midrule
\multicolumn{6}{@{}l}{\textit{NIP-out}} \\
Base & 0.023 & 0.039 & 0.050 & 0.093 & 0.250 \\
MARI-B & 0.012 & 0.015 & 0.014 & 0.061 & 0.270 \\
MARI$^{\dagger\star}$ & 0.013 & 0.046 & 0.084 & 0.133 & 0.317 \\
\textbf{MARI$^\dagger$} & 0.016 & \textbf{0.068} & \textbf{0.114} & \textbf{0.248} & \textbf{0.364} \\
SFT w/o SDM & 0.538 & - & - & - & - \\
\textbf{MARI} & \textbf{0.556} & - & - & - & - \\
\midrule
\multicolumn{6}{@{}l}{\textit{DCP}} \\
Base & 0.462 & 0.451 & 0.448 & 0.421 & 0.516 \\
MARI-B & 0.436 & 0.437 & 0.441 & 0.409 & 0.511 \\
MARI$^{\dagger\star}$ & 0.470 & 0.451 & 0.451 & 0.432 & 0.520 \\
\textbf{MARI$^\dagger$} & \textbf{0.475} & \textbf{0.455} & \textbf{0.470} & \textbf{0.453} & \textbf{0.534} \\
SFT w/o SDM & 0.866 & - & - & - & - \\
\textbf{MARI} & \textbf{0.887} & - & - & - & - \\
\bottomrule
\end{tabular}
\label{table:ablation}
\end{table}

\subsection{Effect of the Number of Retrieved Memories ($K$)}
\begin{figure}[htbp]
\centering
\includegraphics[width=0.8\linewidth]{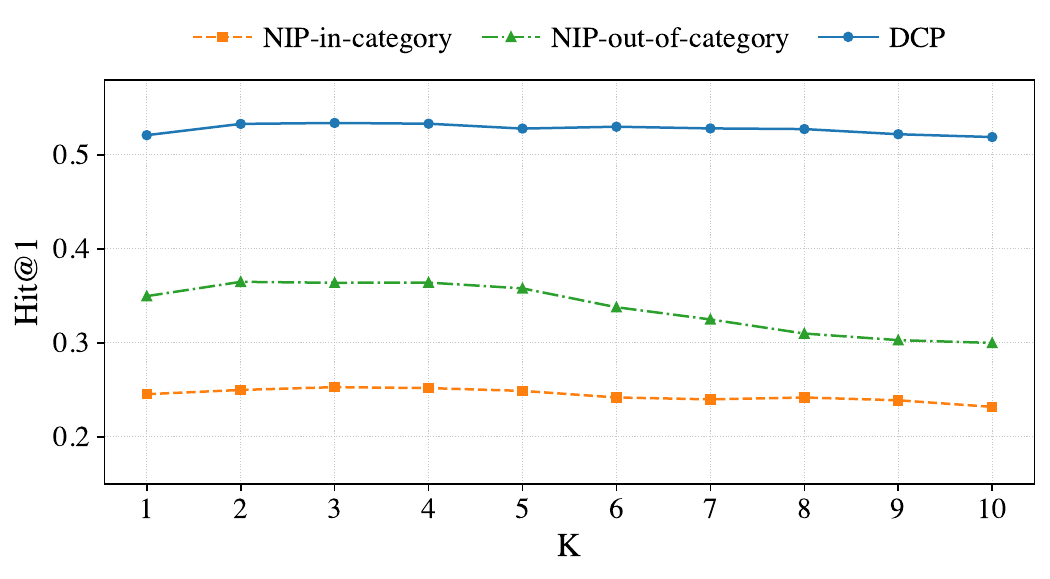}
\caption{MARI$^{\dagger}$'s Hit@1 performance across different tasks as $K$ increases.}
\label{figure:k}
\end{figure}

We investigate the influence of the retrieved memory count ($K$) on MARI$^{\dagger}$'s performance, using DeepSeek-R1 as the underlying recommendation model. Results in Figure~\ref{figure:k} show a consistent trend: performance initially improves with $K$, peaks around $K=3$ or 4, and then declines. Sensitivity to $K$ varies by task. The NIP-out task is most sensitive, as larger $K$ introduces noise that impairs cross-category reasoning. Conversely, NIP-in and DCP performance remains stable across a broader $K$ range, demonstrating robustness to context volume. We set $K=3$ as the default, achieving a balance of strong performance across tasks while minimizing noise from excessive context.

\subsection{Memory Error Analysis}
To evaluate the performance lower bound of MARI under degraded SDM quality, we simulate two error scenarios: (1) \textit{error retrieval} (lowest-similarity memories) and (2) \textit{error memory} (random same-category SDMs from other users). Table~\ref{table:erro} reveals three key insights:

\textbf{(1) Structural priors defy noise.} On NIP-in, error variants consistently outperform the Base LLM (e.g., boosting 4B from 0.075 to 0.094). This suggests even inaccurate SDMs provide valuable structural signals guiding attention better than raw behaviors.

\textbf{(2) Robustness scales with capacity.} On NIP-out, small models (4B/14B) degrade slightly, whereas larger models ($\ge$20B) remain robust and even improve (e.g., R1 jumps to 0.307). This indicates larger LLMs possess sufficient intrinsic capacity to filter noise.

\textbf{(3) Factual sensitivity in DCP.} Performance dips modestly but stays competitive, as DCP inherently relies on precise preference matching, making it more sensitive to factual inaccuracies.

Overall, while LLMs exhibit commendable robustness, high-quality memories remain indispensable for maximizing complex reasoning performance.

\begin{table}[htbp] 
\centering
\caption{Memory Error Analysis. \textit{Err. Ret.}: Lowest-similarity memories. \textit{Err. Mem.}: Random same-category SDMs from other users.}
\label{tab:robustness}
\footnotesize
\setlength{\tabcolsep}{3.5pt} 
\begin{tabular}{@{} l *{5}{c} @{}}
\toprule
\textbf{Variant} & \textbf{Q4B} & \textbf{Q14B} & \textbf{Q32B} & \textbf{GPT-OSS} & \textbf{R1} \\
\midrule
\multicolumn{6}{@{}l}{\textit{NIP-in}} \\
Base & 0.075 & 0.100 & 0.114 & 0.148 & 0.196 \\
Err. Ret. & 0.079 & 0.124 & 0.124 & 0.158 & 0.230 \\
Err. Mem. & 0.094 & 0.120 & 0.117 & 0.151 & 0.223 \\
\textbf{MARI$^\dagger$} & \textbf{0.128} & \textbf{0.154} & \textbf{0.174} & \textbf{0.196} & \textbf{0.253} \\
\midrule
\multicolumn{6}{@{}l}{\textit{NIP-out}} \\
Base & 0.023 & 0.039 & 0.050 & 0.093 & 0.250 \\
Err. Ret. & 0.013 & 0.038 & 0.080 & 0.129 & 0.307 \\
Err. Mem. & 0.012 & 0.032 & 0.062 & 0.118 & 0.301 \\ 
\textbf{MARI$^\dagger$} & \textbf{0.016} & \textbf{0.068} & \textbf{0.114} & \textbf{0.248} & \textbf{0.364} \\
\midrule
\multicolumn{6}{@{}l}{\textit{DCP}} \\
Base & 0.462 & 0.451 & 0.448 & 0.421 & 0.516 \\
Err. Ret. & 0.437 & 0.398 & 0.432 & 0.428 & 0.495 \\
Err. Mem. & 0.443 & 0.414 & 0.438 & 0.405 & 0.494 \\
\textbf{MARI$^\dagger$} & \textbf{0.475} & \textbf{0.455} & \textbf{0.470} & \textbf{0.453} & \textbf{0.534} \\
\bottomrule
\end{tabular}
\label{table:erro}
\end{table}

\section{Conclusion}
We propose MARI, a reasoning-aware framework that explicitly models the decision logic behind heterogeneous user interactions. Via Post-Hoc Decision Distillation, MARI constructs a Decision Memory Bank of structured rationales capturing user goals and trade-offs. Retrieving these memories grounds LLM recommendations in explicit historical evidence, ensuring both interpretability and efficiency. Extensive offline experiments confirm MARI's superiority in accuracy and real-world engagement, offering a practical path toward transparent, rationale-driven recommender systems. Future work will explore advanced SDM retrieval strategies and adaptive memory updates for dynamic environments, and quantify the specific contribution of individual SDM dimensions to recommendation quality and interpretability.

\bibliographystyle{IEEEtran}
\bibliography{references}

@inproceedings{chaney2015probabilistic,
  title={A probabilistic model for using social networks in personalized item recommendation},
  author={Chaney, Allison JB and Blei, David M and Eliassi-Rad, Tina},
  booktitle={Proceedings of the 9th ACM Conference on Recommender Systems},
  pages={43--50},
  year={2015}
}

@inproceedings{he2017neural,
  title={Neural collaborative filtering},
  author={He, Xiangnan and Liao, Lizi and Zhang, Hanwang and Nie, Liqiang and Hu, Xia and Chua, Tat-Seng},
  booktitle={Proceedings of the 26th international conference on world wide web},
  pages={173--182},
  year={2017}
}

@inproceedings{kim2023task,
  title={Task relation-aware continual user representation learning},
  author={Kim, Sein and Lee, Namkyeong and Kim, Donghyun and Yang, Minchul and Park, Chanyoung},
  booktitle={Proceedings of the 29th ACM SIGKDD Conference on Knowledge Discovery and Data Mining},
  pages={1107--1119},
  year={2023}
}

@article{mnih2007probabilistic,
  title={Probabilistic matrix factorization},
  author={Mnih, Andriy and Salakhutdinov, Russ R},
  journal={Advances in neural information processing systems},
  volume={20},
  year={2007}
}

@inproceedings{kang2018self,
  title={Self-attentive sequential recommendation},
  author={Kang, Wang-Cheng and McAuley, Julian},
  booktitle={2018 IEEE international conference on data mining (ICDM)},
  pages={197--206},
  year={2018},
  organization={IEEE}
}

@inproceedings{yang2022multi,
  title={Multi-behavior hypergraph-enhanced transformer for sequential recommendation},
  author={Yang, Yuhao and Huang, Chao and Xia, Lianghao and Liang, Yuxuan and Yu, Yanwei and Li, Chenliang},
  booktitle={Proceedings of the 28th ACM SIGKDD conference on knowledge discovery and data mining},
  pages={2263--2274},
  year={2022}
}

@inproceedings{sun2019bert4rec,
  title={BERT4Rec: Sequential recommendation with bidirectional encoder representations from transformer},
  author={Sun, Fei and Liu, Jun and Wu, Jian and Pei, Changhua and Lin, Xiao and Ou, Wenwu and Jiang, Peng},
  booktitle={Proceedings of the 28th ACM international conference on information and knowledge management},
  pages={1441--1450},
  year={2019}
}

@inproceedings{ren2024representation,
  title={Representation learning with large language models for recommendation},
  author={Ren, Xubin and Wei, Wei and Xia, Lianghao and Su, Lixin and Cheng, Suqi and Wang, Junfeng and Yin, Dawei and Huang, Chao},
  booktitle={Proceedings of the ACM web conference 2024},
  pages={3464--3475},
  year={2024}
}

@inproceedings{zhang2024notellm,
  title={Notellm: A retrievable large language model for note recommendation},
  author={Zhang, Chao and Wu, Shiwei and Zhang, Haoxin and Xu, Tong and Gao, Yan and Hu, Yao and Chen, Enhong},
  booktitle={Companion Proceedings of the ACM Web Conference 2024},
  pages={170--179},
  year={2024}
}

@inproceedings{geng2024breaking,
  title={Breaking the length barrier: Llm-enhanced CTR prediction in long textual user behaviors},
  author={Geng, Binzong and Huan, Zhaoxin and Zhang, Xiaolu and He, Yong and Zhang, Liang and Yuan, Fajie and Zhou, Jun and Mo, Linjian},
  booktitle={Proceedings of the 47th International ACM SIGIR Conference on Research and Development in Information Retrieval},
  pages={2311--2315},
  year={2024}
}

@inproceedings{bao2023tallrec,
  title={Tallrec: An effective and efficient tuning framework to align large language model with recommendation},
  author={Bao, Keqin and Zhang, Jizhi and Zhang, Yang and Wang, Wenjie and Feng, Fuli and He, Xiangnan},
  booktitle={Proceedings of the 17th ACM conference on recommender systems},
  pages={1007--1014},
  year={2023}
}

@inproceedings{kim2024large,
  title={Large language models meet collaborative filtering: An efficient all-round llm-based recommender system},
  author={Kim, Sein and Kang, Hongseok and Choi, Seungyoon and Kim, Donghyun and Yang, Minchul and Park, Chanyoung},
  booktitle={Proceedings of the 30th ACM SIGKDD Conference on Knowledge Discovery and Data Mining},
  pages={1395--1406},
  year={2024}
}

@inproceedings{kim2025lost,
  title={Lost in Sequence: Do Large Language Models Understand Sequential Recommendation?},
  author={Kim, Sein and Kang, Hongseok and Kim, Kibum and Kim, Jiwan and Kim, Donghyun and Yang, Minchul and Oh, Kwangjin and McAuley, Julian and Park, Chanyoung},
  booktitle={Proceedings of the 31st ACM SIGKDD Conference on Knowledge Discovery and Data Mining V. 2},
  pages={1160--1171},
  year={2025}
}

@inproceedings{hou2024large,
  title={Large language models are zero-shot rankers for recommender systems},
  author={Hou, Yupeng and Zhang, Junjie and Lin, Zihan and Lu, Hongyu and Xie, Ruobing and McAuley, Julian and Zhao, Wayne Xin},
  booktitle={European Conference on Information Retrieval},
  pages={364--381},
  year={2024},
  organization={Springer}
}

@inproceedings{ning2025user,
  title={User-llm: Efficient llm contextualization with user embeddings},
  author={Ning, Lin and Liu, Luyang and Wu, Jiaxing and Wu, Neo and Berlowitz, Devora and Prakash, Sushant and Green, Bradley and O'Banion, Shawn and Xie, Jun},
  booktitle={Companion Proceedings of the ACM on Web Conference 2025},
  pages={1219--1223},
  year={2025}
}

@inproceedings{zheng2024harnessing,
  title={Harnessing large language models for text-rich sequential recommendation},
  author={Zheng, Zhi and Chao, Wenshuo and Qiu, Zhaopeng and Zhu, Hengshu and Xiong, Hui},
  booktitle={Proceedings of the ACM Web Conference 2024},
  pages={3207--3216},
  year={2024}
}

@article{kim2408review,
  title={Review-driven personalized preference reasoning with large language models for recommendation. CoRR, abs/2408.06276, 2024. doi: 10.48550},
  author={Kim, Jieyong and Kim, Hyunseo and Cho, Hyunjin and Kang, SeongKu and Chang, Buru and Yeo, Jinyoung and Lee, Dongha},
  journal={arXiv preprint ARXIV.2408.06276},
  year={2024}
}

@inproceedings{wu2024coral,
  title={Coral: collaborative retrieval-augmented large language models improve long-tail recommendation},
  author={Wu, Junda and Chang, Cheng-Chun and Yu, Tong and He, Zhankui and Wang, Jianing and Hou, Yupeng and McAuley, Julian},
  booktitle={Proceedings of the 30th ACM SIGKDD Conference on Knowledge Discovery and Data Mining},
  pages={3391--3401},
  year={2024}
}

@article{gao2019learning,
  title={Learning to recommend with multiple cascading behaviors},
  author={Gao, Chen and He, Xiangnan and Gan, Dahua and Chen, Xiangning and Feng, Fuli and Li, Yong and Chua, Tat-Seng and Yao, Lina and Song, Yang and Jin, Depeng},
  journal={IEEE transactions on knowledge and data engineering},
  volume={33},
  number={6},
  pages={2588--2601},
  year={2019},
  publisher={IEEE}
}

@inproceedings{gu2022self,
  title={Self-supervised Graph Neural Networks for Multi-behavior Recommendation.},
  author={Gu, Shuyun and Wang, Xiao and Shi, Chuan and Xiao, Ding},
  booktitle={IJCAI},
  pages={2052--2058},
  year={2022}
}

@article{qwen3,
    title={Qwen3 Technical Report}, 
    author={An Yang and Anfeng Li and Baosong Yang and Beichen Zhang and Binyuan Hui and Bo Zheng and Bowen Yu and Chang Gao and Chengen Huang and Chenxu Lv and Chujie Zheng and Dayiheng Liu and Fan Zhou and Fei Huang and Feng Hu and Hao Ge and Haoran Wei and Huan Lin and Jialong Tang and Jian Yang and Jianhong Tu and Jianwei Zhang and Jianxin Yang and Jiaxi Yang and Jing Zhou and Jingren Zhou and Junyang Lin and Kai Dang and Keqin Bao and Kexin Yang and Le Yu and Lianghao Deng and Mei Li and Mingfeng Xue and Mingze Li and Pei Zhang and Peng Wang and Qin Zhu and Rui Men and Ruize Gao and Shixuan Liu and Shuang Luo and Tianhao Li and Tianyi Tang and Wenbiao Yin and Xingzhang Ren and Xinyu Wang and Xinyu Zhang and Xuancheng Ren and Yang Fan and Yang Su and Yichang Zhang and Yinger Zhang and Yu Wan and Yuqiong Liu and Zekun Wang and Zeyu Cui and Zhenru Zhang and Zhipeng Zhou and Zihan Qiu},
    journal = {arXiv preprint arXiv:2505.09388},
    year={2025}
}

@article{qwen3embedding,
  title={Qwen3 Embedding: Advancing Text Embedding and Reranking Through Foundation Models},
  author={Zhang, Yanzhao and Li, Mingxin and Long, Dingkun and Zhang, Xin and Lin, Huan and Yang, Baosong and Xie, Pengjun and Yang, An and Liu, Dayiheng and Lin, Junyang and Huang, Fei and Zhou, Jingren},
  journal={arXiv preprint arXiv:2506.05176},
  year={2025}
}

@misc{deepseekai2025deepseekr1incentivizingreasoningcapability,
      title={DeepSeek-R1: Incentivizing Reasoning Capability in LLMs via Reinforcement Learning}, 
      author={DeepSeek-AI},
      year={2025},
      eprint={2501.12948},
      archivePrefix={arXiv},
      primaryClass={cs.CL},
      url={https://arxiv.org/abs/2501.12948}, 
}

@misc{openai2025gptoss120bgptoss20bmodel,
      title={gpt-oss-120b \& gpt-oss-20b Model Card}, 
      author={OpenAI},
      year={2025},
      eprint={2508.10925},
      archivePrefix={arXiv},
      primaryClass={cs.CL},
      url={https://arxiv.org/abs/2508.10925}, 
}

@article{hou2024bridging,
  title={Bridging language and items for retrieval and recommendation},
  author={Hou, Yupeng and Li, Jiacheng and He, Zhankui and Yan, An and Chen, Xiusi and McAuley, Julian},
  journal={arXiv preprint arXiv:2403.03952},
  year={2024}
}

@inproceedings{zhao2015improving,
  title={Improving user topic interest profiles by behavior factorization},
  author={Zhao, Zhe and Cheng, Zhiyuan and Hong, Lichan and Chi, Ed H},
  booktitle={Proceedings of the 24th international conference on world wide web},
  pages={1406--1416},
  year={2015}
}

@inproceedings{ding2018improving,
  title={Improving Implicit Recommender Systems with View Data.},
  author={Ding, Jingtao and Yu, Guanghui and He, Xiangnan and Quan, Yuhan and Li, Yong and Chua, Tat-Seng and Jin, Depeng and Yu, Jiajie},
  booktitle={IJCAI},
  pages={3343--3349},
  year={2018}
}

@article{guo2017resolving,
  title={Resolving data sparsity by multi-type auxiliary implicit feedback for recommender systems},
  author={Guo, Guibing and Qiu, Huihuai and Tan, Zhenhua and Liu, Yuan and Ma, Jing and Wang, Xingwei},
  journal={Knowledge-Based Systems},
  volume={138},
  pages={202--207},
  year={2017},
  publisher={Elsevier}
}

@inproceedings{guo2019buying,
  title={Buying or browsing?: Predicting real-time purchasing intent using attention-based deep network with multiple behavior},
  author={Guo, Long and Hua, Lifeng and Jia, Rongfei and Zhao, Binqiang and Wang, Xiaobo and Cui, Bin},
  booktitle={Proceedings of the 25th ACM SIGKDD international conference on knowledge discovery \& data mining},
  pages={1984--1992},
  year={2019}
}

@inproceedings{xia2020multiplex,
  title={Multiplex behavioral relation learning for recommendation via memory augmented transformer network},
  author={Xia, Lianghao and Huang, Chao and Xu, Yong and Dai, Peng and Zhang, Bo and Bo, Liefeng},
  booktitle={Proceedings of the 43rd international ACM SIGIR conference on research and development in information retrieval},
  pages={2397--2406},
  year={2020}
}

@inproceedings{cheng2023multi,
  title={Multi-behavior recommendation with cascading graph convolution networks},
  author={Cheng, Zhiyong and Han, Sai and Liu, Fan and Zhu, Lei and Gao, Zan and Peng, Yuxin},
  booktitle={Proceedings of the ACM Web Conference 2023},
  pages={1181--1189},
  year={2023}
}

@article{ning2025retrieval,
  title={Retrieval-Augmented Purifier for Robust LLM-Empowered Recommendation},
  author={Ning, Liangbo and Fan, Wenqi and Li, Qing},
  journal={arXiv preprint arXiv:2504.02458},
  year={2025}
}

@inproceedings{lee2024mule,
  title={Mule: Multi-grained graph learning for multi-behavior recommendation},
  author={Lee, Seunghan and Ko, Geonwoo and Song, Hyun-Je and Jung, Jinhong},
  booktitle={Proceedings of the 33rd ACM international conference on information and knowledge management},
  pages={1163--1173},
  year={2024}
}

@inproceedings{zhai2025combinatorial,
  title={Combinatorial optimization perspective based framework for multi-behavior recommendation},
  author={Zhai, Chenhao and Meng, Chang and Yang, Yu and Zhang, Kexin and Zhao, Xuhao and Li, Xiu},
  booktitle={Proceedings of the 31st ACM SIGKDD Conference on Knowledge Discovery and Data Mining V. 1},
  pages={1891--1902},
  year={2025}
}

@inproceedings{zhang2025hierarchical,
  title={Hierarchical Graph Information Bottleneck for Multi-Behavior Recommendation},
  author={Zhang, Hengyu and Shen, Chunxu and Sun, Xiangguo and Tan, Jie and Tan, Yanchao and Rong, Yu and Cheng, Hong and Yi, Lingling},
  booktitle={Proceedings of the Nineteenth ACM Conference on Recommender Systems},
  pages={155--164},
  year={2025}
}

@article{lewis2020retrieval,
  title={Retrieval-augmented generation for knowledge-intensive nlp tasks},
  author={Lewis, Patrick and Perez, Ethan and Piktus, Aleksandra and Petroni, Fabio and Karpukhin, Vladimir and Goyal, Naman and K{\"u}ttler, Heinrich and Lewis, Mike and Yih, Wen-tau and Rockt{\"a}schel, Tim and others},
  journal={Advances in neural information processing systems},
  volume={33},
  pages={9459--9474},
  year={2020}
}

@inproceedings{magister2023teaching,
  title={Teaching small language models to reason},
  author={Magister, Lucie Charlotte and Mallinson, Jonathan and Adamek, Jakub and Malmi, Eric and Severyn, Aliaksei},
  booktitle={Proceedings of the 61st Annual Meeting of the Association for Computational Linguistics (Volume 2: Short Papers)},
  pages={1773--1781},
  year={2023}
}

@article{wei2022chain,
  title={Chain-of-thought prompting elicits reasoning in large language models},
  author={Wei, Jason and Wang, Xuezhi and Schuurmans, Dale and Bosma, Maarten and Xia, Fei and Chi, Ed and Le, Quoc V and Zhou, Denny and others},
  journal={Advances in neural information processing systems},
  volume={35},
  pages={24824--24837},
  year={2022}
}

\appendix

\subsection{Item Metadata Example}
\label{app:item}

\begin{center}
\begin{tcolorbox}[
    width=0.98\columnwidth, 
    colback=black!3!white,  
    colframe=black!60!white,
    title=\textbf{Example: Raw Item Profile in the Dataset},
    fonttitle=\small\bfseries, 
    fontupper=\footnotesize,  
    boxrule=0.4pt, arc=1pt,
    left=5pt, right=5pt, top=2pt, bottom=3pt,
    before skip=0.5em, after skip=0.5em
]
\noindent\textbf{Item ID:} 12345678 \hfill \textbf{Category:} Four-Wheel Strollers\\[0.3em]
\noindent\textbf{Brand:} Phoenix \hfill \textbf{Price:} 2000.0\\[0.4em]
\noindent\textbf{Title:} Phoenix 0-3 Years Reclining \& Seating Baby Stroller\\[0.3em] 
\noindent\textbf{Attributes:} Color: Camel [Premium Edition] - Upgraded polished aluminum frame, electroplated EVA silent wheels, includes bonus cartoon seat cushion.\\[0.4em]
\noindent\textbf{Engagement Statistics:}\\
\hspace*{0.5em} Stock: 8,600 \hfill Favorites: 1,900 \hfill Total Sales: 2,700
\end{tcolorbox}
\end{center}

\subsection{Candidate Set Example}
\label{app:candi}
\begin{center}
\begin{tcolorbox}[
    width=0.98\columnwidth, 
    colback=black!3!white,  
    colframe=black!60!white,
    title=\textbf{Example: Candidate Set},
    fonttitle=\small\bfseries, 
    boxrule=0.4pt, arc=1pt,
    left=2pt, right=2pt, top=2pt, bottom=2pt,
    before skip=0.5em, after skip=0.5em
]
\begin{lstlisting}[
    basicstyle=\scriptsize\ttfamily, % 7pt 等宽字体
    breaklines=true,                 % 核心：允许长行自动换行
    columns=fullflexible,
    keepspaces=true,
    showstringspaces=false,
    aboveskip=0pt, belowskip=0pt
]
[
  {"item_title": "PiP Studio Unisex Dutch Imported Mug", "attribute": "Color: Orange", "brand": "PiP Studio", "price": "105.0", "stock": 620, "favorites_count": 2410, "sales_count": 853, "item_id": "578462125xxxx"},
  {"item_title": "PiP Studio Unisex Dutch Imported Mug", "attribute": "Color: Chocolate Brown", "brand": "PiP Studio", "price": "118.0", "stock": 582, "favorites_count": 2410, "sales_count": 878, "item_id": "578114651xxxx"},
  {"item_title": "PiP Studio Unisex Dutch Imported Mug", "attribute": "Color: Sky Blue", "brand": "PiP Studio", "price": "105.0", "stock": 620, "favorites_count": 2410, "sales_count": 853, "item_id": "578114651xxxx"}
]
\end{lstlisting}
\end{tcolorbox}
\end{center}

\subsection{Structured Decision Memory Example}
\label{app:SDM}

\begin{center}
\begin{tcolorbox}[
    width=0.9\columnwidth, 
    colback=black!3!white,  
    colframe=black!60!white,
    title=\textbf{Example: Structured Decision Memory},
    fonttitle=\small\bfseries, 
    fontupper=\scriptsize,  
    boxrule=0.4pt, arc=1pt,
    boxsep=1pt,             
    left=2pt, right=2pt, top=1pt, bottom=1pt, 
    before skip=2pt, after skip=2pt, 
    before upper={\linespread{0.85}\selectfont\setlength{\parskip}{0pt}\setlength{\parindent}{0pt}} 
]
\noindent\textbf{Profile:} Brand Loyalist \hfill \textbf{Efficiency:} Considered\vspace{2pt} 

\noindent\textbf{Intent Reasoning:} 
User exhibited highly goal-directed behavior for women's comfort loungewear. On Oct 1 (15:50), searched ``Fenteng cotton loungewear'', inspected attributes (XXL, beige), added to cart, and repeatedly viewed media to validate comfort. At 15:52, briefly explored an alternative (``Hannaguo modal loungewear set'') but returned to Fenteng within 1 min, reinforcing brand trust. After a cooling-off period, replayed video/images on Oct 7 (09:38) for final confirmation, purchasing at 125 (orig. 318) at 09:40. Core drivers: brand reliability and material safety; alternative exploration merely reinforced the initial choice.\vspace{2pt} 

\noindent\textbf{Preference Factors:}
\begin{itemize}[leftmargin=*, nosep, topsep=0pt, partopsep=0pt, itemsep=0pt, parsep=0pt] 
    \item \textbf{Brand}: Fenteng -- Consistent focus from search to purchase with repeated interactions. \textit{(High)}
    \item \textbf{Material}: Cotton -- Explicitly required; no non-cotton alternatives considered. \textit{(High)}
    \item \textbf{Price Range}: 100-200 -- Final payment (125) falls in range; accepted discount. \textit{(Medium)}
\end{itemize}\vspace{1pt} 

\noindent\textbf{Avoidance Factors:}
\begin{itemize}[leftmargin=*, nosep, topsep=0pt, partopsep=0pt, itemsep=0pt, parsep=0pt] 
    \item \textbf{Material}: Modal -- Briefly viewed Hannaguo modal product but abandoned it in favor of cotton option \textit{(Medium)}
\end{itemize}
\end{tcolorbox}
\end{center}

\subsection{Prompt Templates}
\label{app:prompt}
\centering
\begin{tcolorbox}[
    width=\linewidth, 
    colback=black!5!white,
    colframe=black!70!white,
    title=\textbf{Prompt Template for MARI}, 
    fonttitle=\small\bfseries, 
    fontupper=\scriptsize\ttfamily, 
    boxrule=0.5pt, arc=2pt,
    left=3pt, right=3pt, top=3pt, bottom=3pt,
    before upper={\linespread{0.9}\selectfont\setlength{\parskip}{2pt}\setlength{\parindent}{0pt}} 
]
\textbf{\# Role \& Task}\par 
You are a senior e-commerce shopping advisor. Given the user's [Historical Decisions], [Recent Behaviors], and [Candidate Items], predict the single item the user is most likely to purchase next.\vspace{2pt}

\textbf{\# Decision Logic (in order of priority)}\par
1. \textbf{Intent Recognition}: Analyze commonalities across recent behaviors and decisions (e.g., brand preference, price range, urgent needs).\par
2. \textbf{Logical Judgment}:\par
\hspace*{1em}- If \textbf{[Candidate Items]} is non-empty: Select the single item from the pool that best aligns with the user's recent behavior and exhibits the strongest purchase intent.\par
\hspace*{1em}- If \textbf{[Candidate Items]} is empty: Revisit [Recent Behaviors] and select the single item with the strongest recent purchase intent.\vspace{2pt}

\textbf{\# Output Format (Strict JSON)}\par
\{\par
\hspace*{1em}"reasoning": "A logical and coherent explanation",\par 
\hspace*{1em}"item\_title": "Full name of the predicted item",\par
\hspace*{1em}"item\_id": "Unique ID of the predicted item"\par
\}\vspace{2pt}

\textbf{\# Runtime Input Variables}\par 
Historical Decisions: \{Historical\_Decisions\}\par
Recent Behaviors: \{Recent\_Behaviors\}\par
Candidate Items: \{Candidate\_Items\}
\end{tcolorbox}

\end{document}